\pdfoutput=1

\documentclass[11pt]{article}

\usepackage{emnlp2021}
\usepackage{hyphenat}
\usepackage{times}
\usepackage{latexsym}
\usepackage[T1]{fontenc}
\usepackage[utf8]{inputenc}
\usepackage{microtype}
\usepackage{graphicx}       
\usepackage{dblfloatfix}    
\usepackage{tabularx}       
\usepackage{tablefootnote}  

\usepackage{soul}
\newcommand{\cul}[2]{\setulcolor{#1}\ul{#2}}  
\usepackage{todonotes}
\usepackage{color}
\newcommand{\thightbox}[1]{{\setlength{\fboxsep}{1.33pt}\framebox{#1}}}
\definecolor{olivine}{rgb}{0.6, 0.73, 0.45}

\title{BioLORD: Learning Ontological Representations from Definitions \\
for Biomedical Concepts and their Textual Descriptions
}

\author{François Remy \and Kris Demuynck \and Thomas Demeester \\
  The Internet and Data Science Lab (IDLab) \\
  Ghent University (UGent) - Imec Belgium \\
  \texttt{francois.remy@ugent.be} \\}

\begin{document}
\maketitle
\begin{abstract}
\vspace{-0.19cm}
\makebox[1.0\linewidth][s]{This work introduces BioLORD, a new pre-}
\makebox[1.0\linewidth][s]{training strategy for producing meaningful rep-}
\makebox[1.0\linewidth][s]{resentations for clinical sentences and bio-}
\makebox[1.0\linewidth][s]{medical concepts. State-of-the-art methodolo-}
\makebox[1.0\linewidth][s]{gies operate by maximizing the similarity in} \makebox[1.0\linewidth][s]{representation of names referring to the same}
\makebox[1.0\linewidth][s]{concept, and preventing collapse through con-}
\makebox[1.0\linewidth][s]{trastive learning. However, because \mbox{biomedical}}
\makebox[1.0\linewidth][s]{names are not always self-explanatory, it some-}
\makebox[1.0\linewidth][s]{times results in non-semantic representations.}
\makebox[1.0\linewidth][s]{\mbox{BioLORD} overcomes this issue by ground-}
\makebox[1.0\linewidth][s]{ing concept representations using definitions,}
\makebox[1.0\linewidth][s]{as well as short descriptions derived from a}
\makebox[1.0\linewidth][s]{multi-relational knowledge graph consisting of}
\makebox[1.0\linewidth][s]{biomedical ontologies. Thanks to this ground-}
\makebox[1.0\linewidth][s]{ing, our model produces more semantic concept}
\makebox[1.0\linewidth][s]{representations that match more closely the hi-}
\makebox[1.0\linewidth][s]{erarchical structure of ontologies. BioLORD}
\makebox[1.0\linewidth][s]{establishes a new state of the art for text simi-}
\makebox[1.0\linewidth][s]{larity on both clinical sentences (MedSTS) and}
\makebox[1.0\linewidth][l]{biomedical concepts (MayoSRS).}
\end{abstract}

\begin{figure*}[t]
\begin{center}
\includegraphics[width=0.9999\linewidth]{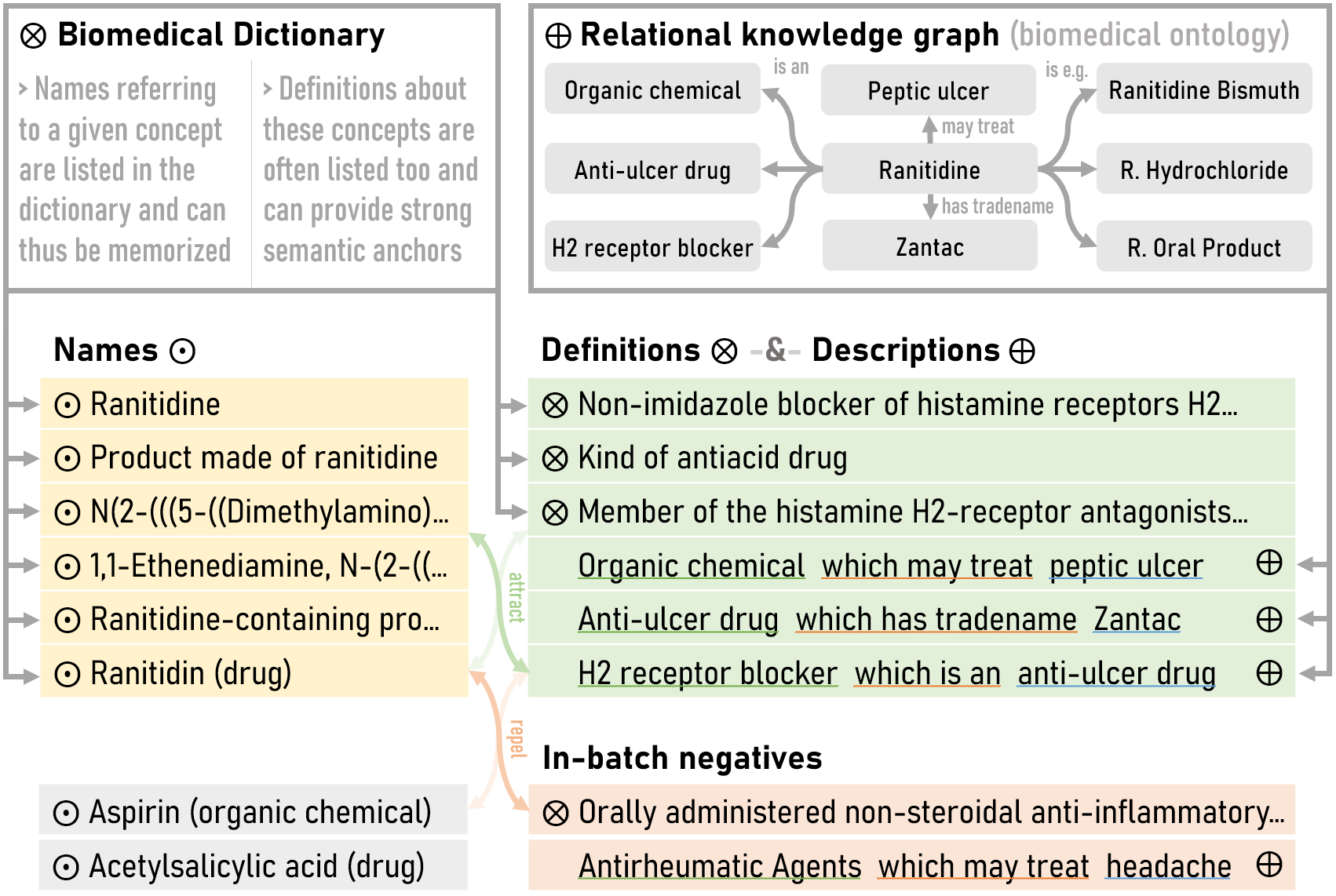}    
\end{center}
\caption{BioLORD aims to bring the representation of biomedical concept \textbf{names} ($\odot$) and their \textbf{definitions} ($\otimes$)
closer to each other, to ground the name representations with knowledge from the definitions. This is illustrated for the \textit{\href{https://uts.nlm.nih.gov/uts/umls/concept/C0034665}{Ranitidine}}
and \textit{\href{https://uts.nlm.nih.gov/uts/umls/concept/C0004057}{Aspirin}} concepts from UMLS.
Knowledge from the ontology's relational knowledge graph is injected by extending the set of known definitions with automatically generated \textbf{descriptions} ($\oplus$). 
Each such description pairs a \cul{olivine}{more generic concept} with one  \cul{orange}{relationship} (of the described concept) and its \cul{blue}{related concept}, thereby setting the described concept apart from the more generic one.
Contrastive learning is applied to \textit{attract} the representations of compatible pairs ($\odot$, $\otimes$ or $\oplus$) and \textit{repel} incompatible ones (obtained as \textbf{in-batch negatives}).
}
\label{fig:main}
\end{figure*}

\vspace{-0.2cm}
\section{Introduction}
\label{section:Introduction}

Natural language processing models are well positioned to support healthcare providers by automatically extracting and synthesizing relevant information from clinical notes.
For this, we introduce BioLORD, a pre-training strategy for end-to-end biomedical information extraction, 
capable of producing meaningful representations for biomedical terms and clinical sentences simultaneously. 

This is achieved through the continued pre-training of an existing sentence embedding model, using contrastive learning and 
pairs consisting of the names and definitions of a given biomedical concept (see Fig.~\ref{fig:main}). 
This design choice proved crucial for the effectiveness of BioLORD, as it enables the transfer of knowledge from the definitions 
to the representation of biomedical names, thereby overcoming limitations of existing works (see §\ref{section:ChallengesSOTA}) through a more effective usage of the knowledge contained in biomedical ontologies (see §\ref{section:Ontologies}).

Indeed, to improve coverage and diversity, we supplemented definitions with textual descriptions 
generated from the numerous concept-to-concept relationships contained in biomedical ontologies.

Our key contributions are 
\thightbox{1} a versatile training strategy using dictionaries and knowledge graphs to create highly semantic representations for the key phrases of a domain, 
\thightbox{2} an associated BioLORD model trained on the biomedical domain, 
\thightbox{3} an extensive evaluation (§\ref{section:Evaluation}) demonstrating its ability to provide semantic representations usable in a broad range of information extraction scenarios, including a new state of the art for Biomedical Concept Representation and Clinical Sentence Similarity, and 
\thightbox{4} an in-depth analysis of the strengths and weaknesses of our proposed approach (§\ref{section:Discussion}).

\section{Related Work}
\label{section:RelatedWork}

Let us first consider how prior works attempted to address the biomedical domain's usage of a large, specialized, and often opaque vocabulary (e.g., \textit{PAPA syndrome}\footnote{a hereditary inflammatory disorder affecting the skin} or \textit{cat scratch disease}\footnote{a bacterial skin infection caused by Bartonella Henselae}).

\subsection{Biomedical ontologies} 
\label{section:Ontologies} 

To condense this lexical knowledge in digital form, medical practitioners developed semi-structured concept hierarchies called biomedical ontologies, merging a dictionary and a knowledge graph.

\textbf{SnomedCT} (Systematized Nomenclature of Medicine and Clinical Terms) is one such ontology covering around 700k medical concepts in total and a small set of important relationships between these concepts \citep{SnomedCT}. 

\textbf{UMLS} (Unified Medical Language System) 
bridges several biomedical ontologies to cover
more than 4 million concepts, each with on average 4 listed names \citep{UMLS}. 
UMLS also contains around 90 million labeled concept-to-concept relationships of 900 different types.

\subsection{Contrastive Learning Strategies}
\label{section:RelatedWorkContrastiveLearning}

On the machine learning side, efforts in the tasks of named entity recognition (NER) and normalization (NEL) 
are strongly influenced by the challenges posed by such a large and specialized vocabulary.
In recent years, approaches using ontologies through string-based pattern matching, such as MetaMap \cite{MetaMap}, have been consistently outperformed by newer works relying on constrative learning with Transformers.

\textbf{BioSyn} \citep{sung-etal-2020-biomedical} was the first model to introduce the idea of contrastive learning to produce embeddings of biomedical concepts. It takes existing NEL benchmarks and proposes to use their training sets in a contrastive manner. An encoder model initialized with BioBERT \citep{BioBERT} is trained to produce embeddings for batches of concept names (grouped by pairs referring to the same concept). A contrastive loss is then applied to ensure that the embeddings of synonyms are significantly closer to each other than they are to the other names in the batch, which refer to other concepts. After pre-training, the model can be finetuned for the end task of NEL using cross-entropy training. 

\textbf{SapBERT} \citep{liu-etal-2021-self} was the first large-scale contrastive model to leverage UMLS. Just like BioSyn, it produces embeddings for biomedical concept names, without considering the context they are used in.
But, unlike BioSyn, it is based on PubMedBERT \citep{PubMedBERT} and uses the synonyms defined for concepts in UMLS to form the training pairs. This enables the model to contrast millions of entries, many more than BioSyn. 

\textbf{BIOCOM} \citep{BIOCOM} and \textbf{KRISSBERT} \citep{KRISSBERT} independently extended this approach in a similar way, by noting the need for context-based disambiguation for some entities. For each UMLS concept, sentences mentioning the concept are collected from PubMed articles. These sentences are used as context during training.

\subsection{Challenges with existing models}
\label{section:ChallengesSOTA}

\begin{figure}[t]
    \centering
    \includegraphics[width=\linewidth]{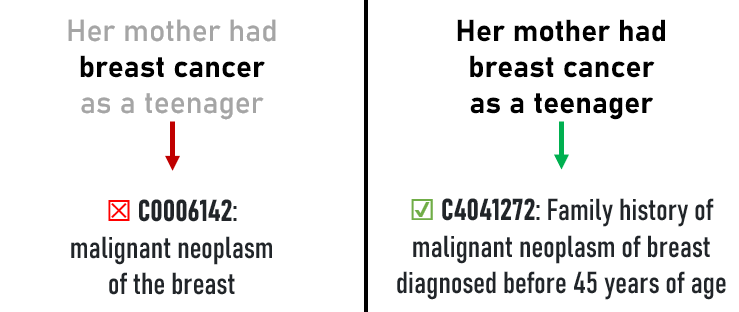}
	    \caption{Concept mapping sometimes requires considering the entire sentence, rather than mentions. 
    }
    \label{fig:map_sentence}
\end{figure}

BIOCOM and KRISSBERT propose to disambiguate mentions of biomedical concepts using contextual information. 
Ambiguous notations requiring context to disambiguate 
can indeed be found in clinical notes. 
However, using these contextual models for inference is only possible after identifying text spans denoting such concepts in the input text. This requires introducing a mention detection model, which comes with its own challenges and errors. Worse, reducing mentions to text spans is not always possible, as concepts are sometimes alluded to in a diffused way (see Fig.~\ref{fig:map_sentence}). 

\begin{figure}[t]
    \centering
    \includegraphics[width=0.9\linewidth]{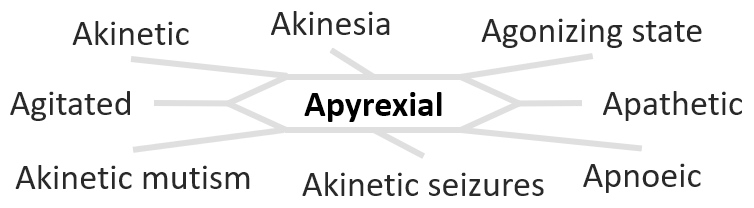}
	    
    \caption{In SapBERT's latent space, none of the nearest neighbors of "apyrexial" (i.e. \textit{fever-free}) happen to share the word's meaning. Instead, the alpha-privative was over-indexed by the model, among other biases.}
    \label{fig:map_belgium}
\end{figure}

However, models which do not use in-context mentions usually learn representations of 
	lower quality than in-context models. By pairing synonyms with a significant word or token overlap with each other, these models isolate concepts containing rare words or tokens early in the training, in a way that is rarely semantic (see Fig.~\ref{fig:map_belgium}). 
	Indeed, the training loss of contrastive models only requires placing all mentions of a particular concept close to each other, but it does not provide strong guarantees about the relative location of different but similar concepts in the latent space. 

While hierarchical relationships from medical ontologies have sometimes been used to produce more meaningful concept embeddings \citep{KRISSBERT}, this is however not sufficient to overcome the issues stated above, because relatedness is not always possible to encode hierarchically.

\section{Pre-training methodology}
\label{section:Methodology}

To produce representations of biomedical concepts that overcome the limitations described above, we modified the way the positive pairs are constructed.
	Like the prior works cited in §\ref{section:RelatedWorkContrastiveLearning}, we start by establishing a list of names for each UMLS concept. 
	However, unlike previous works, we do not use these names directly to form positive pairs. Instead, we construct pairs formed with, on the one side, a randomly selected name for a given concept and, on the other side, a definition or description for that concept
(see Fig.~\ref{fig:main}). 

We hypothesize that a definition or description 
of a given concept provides a more robust semantic anchor for this concept than another of its names. As mentioned before, names in the medical domain can be quite opaque, and do not always offer useful insights into what exactly is being referred to. 
By inducing representational similarity between a concept name and its known definitions,
	we aim to distill their respective knowledge
	into the representations of the concept names themselves. 
	This key idea 
influenced some design choices for our experimental setup, 
	including the choice of 
the data curation process, model initialization, and training procedure (as described in this section). 
	
\subsection{Curating definitions and descriptions}
Around 5\% of the concepts found in UMLS are clarified by one or more definitions. These definitions aim to provide the most relevant pieces of information about a given concept to the practitioners reading them, and we can therefore include them directly in our training set (see Fig.~\ref{fig:main}). 

This is however insufficient, since most concepts have no matching definition in UMLS. Additionally, definitions might not always cover all the relevant aspects of a given concept, and the particular aspects they cover vary from one concept to another. Consequently, pairing concept names and their definitions, alone, cannot be expected to 
produce satisfactory results for all UMLS concepts.

We therefore supplement the definitions already available in UMLS with automatically generated textual descriptions, based on the structured information contained in the ontology and its 90M concept-to-concept relationships.

These concept descriptions are constructed using the following template: “\textit{[more-generic-concept] which [\mbox{has-relationship-with}] [related-concept]}” (e.g. "\textit{drug which may treat headache}").

The replacement for “\textit{[more-generic-concept]}” is randomly sampled among the known names of the ancestors and/or semantic types of that concept, or left blank (i.e. replaced by “\textit{something}”). Most UMLS relationships are already expressed using a verbal form which can be used as-is in the template, but a set of rules was crafted to convert the relationships which were not (usually by prepending “\textit{is}” or “\textit{has}” before their name). Finally, a known name of the related concept is randomly selected to finish forming the description.

The descriptions constructed that way do not always unambiguously refer to a unique concept, but we consider this to be a desirable property 
because it tends to pull closer to each other the concepts
which share characteristics that practitioners found useful to encode as relationships.

\subsection{Pre-training setup}

To maximally leverage the meaning of definitions in contrastive pairs, 
a sentence embedding model trained on 1B positive pairs
(\textbf{STAMB{2}})
	was used as initialization \citep{STAMB2}. 
		As a result, the representations produced for opaque concept names are likely to improve quickly, by drawing insights from the definitions. 
	It is worth noting that the STAMB2 model has seen PubMed titles and abstracts as part of its pre-training; therefore, it already possesses some general understanding of the biomedical domain, albeit a partial one as our evaluation demonstrates.

To continue training this model, we constructed a dataset containing 100 million pairs of concept names and corresponding definitions or descriptions. From all training pairs, 85\% contain a textual description generated using the concept-to-concept relationships from UMLS, and 15\% contain an actual definition (to achieve this, each definition is sampled 50 times, with different concept names whenever this is possible). 
Our analysis found that concepts having multiple definitions are also more likely to appear in clinical notes: oversampling them therefore appears beneficial.
	We release the dataset\footnote{\href{https://huggingface.co/datasets/FremyCompany/BioLORD-DS}{huggingface.co/datasets/FremyCompany/BioLORD-DS}} to make it easier to reproduce our results.

Our model is trained over this large dataset for one epoch, in batches of 64 pairs, using the \mbox{InfoNCE} loss \citep{InfoNCE}. In Appendix \ref{section:ReproducingDetails}, additional details are presented for readers interested in faithfully reproducing our experiments.

\section{Experimental evaluation}
\label{section:Evaluation}

To demonstrate the effectiveness of our definition-based pre-training strategy, we subsequently train and evaluate our model on multiple Semantic Text Similarity (STS) 
tasks. For each benchmark, we compare our results with the state of the art. 
	We also report results for the models BioSyn \citep{sung-etal-2020-biomedical} and SapBERT \citep{liu-etal-2021-self}, since these
	share their encoder architecture (BERT-base) and parameter count (110M) with our model, enabling fair comparison. We also provide results for the base models of BioSyn, SapBERT, and BioLORD: respectively BioBERT \citep{BioBERT}, PubMedBERT \citep{PubMedBERT} and STAMB2 \citep{STAMB2}. When applicable, we use the same finetuning strategy 
for all six models.

Our key evaluation tasks 
are biomedical concept similarity, biomedical concept normalization, and sentence similarity (both in and out of the biomedical domain). In the following paragraphs, 
	we describe 
the datasets and the experimental setup for these tasks, and already refer to the respective results tables.
	We however reserve our thoughts and insights for §\ref{section:Discussion} to enable cross-task comparisons.

\vspace{0.2cm}

\subsection{Biomedical concept similarity}

Given their pre-training strategy, BioLORD models would be expected to produce particularly strong semantic representations of biomedical concepts. To confirm this, we use the cosine distance between their representations to evaluate the degree of similarity between pairs of biomedical concepts. We compare this similarity measure against similarities derived from human judgment. Given the limited size of those datasets, no finetuning is performed.

Following the approach of \citet{KalyanHybridEmbeddings}, we evaluate our model using four benchmarks: MayoSRS, UMNSRS-Similarity, UMNSRS-Relatedness, and EHR-RelB.

\textbf{MayoSRS} \citep{Pakhomov2011-lq} is a dataset formed by 101 clinical term pairs whose relatedness was reported on a 4-point scale by nine medical coders and three physicians. 

\textbf{UMNSRS} \citep{Pakhomov2010-lq} is a pair of datasets, consisting of 725 clinical term pairs whose semantic similarity and relatedness were determined on a continuous scale by 4 clinicians.

\textbf{EHR-RelB} \citep{schulz-etal-2020-biomedical} is a dataset containing 3630 concept pairs sampled from electronic health records, rated for relatedness by 3 doctors.
	
In Table \ref{tab:results1}, we report the Spearman correlation between the similarity scores attributed by a model and the scores attributed by the medical practitioners. Achieving a high correlation on this benchmark indicates that the embedding space defined by the model possesses a latent structure which corresponds well with the human perception of similarity between concepts.

\begin{table*}
    \centering
    \begin{tabularx}{0.925\textwidth} { 
        >{\raggedright\arraybackslash}p{0.125\textwidth} 
        || >{\raggedright\arraybackslash}X
        | >{\raggedright\arraybackslash}X
        || >{\raggedright\arraybackslash}X 
        | >{\raggedright\arraybackslash}X 
        || >{\raggedright\arraybackslash}X 
        | >{\raggedright\arraybackslash}X 
	        }
         & \textbf{BioBERT} & \textbf{BioSyn} & \textbf{PubMedB.} & \textbf{SapBERT} & \textbf{STAMB2} & \textbf{BioLORD}\\
        \textbf{\textsc{MayoSRS}}  & 28.2 & 45.1 & 42.5 & 62.5 & 47.1 & \underline{\textbf{74.7}}†\\
        \textbf{\textsc{Umnsrs}-R} & 25.3 & 39.1 & 25.3 & 47.5 & 43.9 & \textbf{54.4}†\\
        \textbf{\textsc{Umnsrs}-S} & 33.0 & 43.6 & 25.2 & 53.0 & 46.7 & \textbf{56.0}†\\
        \textbf{\textsc{Ehr-RelB}} & 33.8 & 42.5 & 31.7 & 51.7 & 54.1 & \textbf{57.5}†\\
    \end{tabularx}
    \caption{
        Spearman scores obtained on the Biomedical concept similarity benchmarks. Given the small size of these datasets, no finetuning is performed. Cosine similarity between the representations of the concepts performed best in all cases, but Euclidian and Manhattan distances were tested as well. Higher is better. \\
        \footnotesize{
            \\
            †: We also trained a BioLORD model based on PubMedBERT, whose results (69.0, 52.7, 56.7, 51.1) were superior to SapBERT but inferior to the BioLORD model based on STAMB2 which we present in this article. \\
            \\
            Other state-of-the-art models for the tasks: While BioLORD performs best among comparable models, two systems combining text embeddings with graph-based enhancements perform better for some tasks. See works by \citet{KalyanHybridEmbeddings} and \citet{Mao2020-ov}. These techniques could be applied to BioLORD, but this is left as future work.
        }
    }
    \label{tab:results1}
\end{table*}

\begin{table*}
    \centering
    \begin{tabularx}{0.925\textwidth} { 
        >{\raggedright\arraybackslash}p{0.125\textwidth} 
        || >{\raggedright\arraybackslash}X
        | >{\raggedright\arraybackslash}X
        || >{\raggedright\arraybackslash}X 
        | >{\raggedright\arraybackslash}X 
        || >{\raggedright\arraybackslash}X 
        | >{\raggedright\arraybackslash}X 
	        }
        \textbf{\textsc{Sct-L2P}} & \textbf{BioBERT} & \textbf{BioSyn} & \textbf{PubMedB.} & \textbf{SapBERT} & \textbf{STAMB2} & \textbf{BioLORD}\\
        - \textsc{Mrr} & 28.1 & 35.4 & 24.8 & 40.6 & 41.6 & \underline{\textbf{49.9}}\\
        - \textsc{Acc@1} & 20.4 & 25.9 & 17.9 & 29.1 & 31.1 & \underline{\textbf{37.0}}\\
    \end{tabularx}
    \caption{
        Mean-reciprocal rank (MRR) and Top 1 Accuracy (Acc@1) of the similarity mapping of leaf node concepts of Snomed-CT on their parent concepts, after leaving only non-leaf nodes of the Snomed-CT ontology as candidates. 
    }
    \label{tab:results2a}
\end{table*}

\begin{table*}
    \centering
    \begin{tabularx}{0.925\textwidth} { 
        >{\raggedright\arraybackslash}p{0.125\textwidth} 
        || >{\raggedright\arraybackslash}X
        | >{\raggedright\arraybackslash}X
        || >{\raggedright\arraybackslash}X 
        | >{\raggedright\arraybackslash}X 
        || >{\raggedright\arraybackslash}X 
        | >{\raggedright\arraybackslash}X 
	        }
         & \textbf{BioBERT} & \textbf{BioSyn} & \textbf{PubMedB.} & \textbf{SapBERT} & \textbf{STAMB2} & \textbf{BioLORD}\\
        \textbf{\textsc{MedSts}}  & 83.7 & 84.0 & 85.8 & 86.0 & 85.9 & \textbf{86.3}\\
        \textbf{\textsc{Biosses}} & 88.1 & \textbf{92.1} & 91.5 & 89.3 & 90.7 & 84.0\\
        \textbf{\textsc{Sick}}    & 86.8 & 86.7 & 86.3 & 80.3 & \underline{\textbf{90.7}} & 89.3\\
        \textbf{\textsc{Sts}}     & 79.8 & 79.4 & 82.5 & 81.9 & 88.0 & 86.5\\
    \end{tabularx}
    \caption{
        Pearson scores obtained by our model on the various Semantic Text Similarity tasks on which it was finetuned then tested on. Higher is better. We include two SOTA numbers: one for models of equivalent size and one for models of any size. \\
        \footnotesize{
            \\
            Other state-of-the-art models for the tasks: No comparable model surpasses BioLORD in biomedical sentence similarity tasks but, in the the general-purpose sentence similarity benchmark (STS), the DistillBERT model by \citet{sanh2019distilbert} performs better (90.7) than BioLORD. Models performing better than BioLORD for the biomedical tasks exists, but they are significantly larger models: GatorTron by \citet{GatorTron} for MedSTS (89.0), LLM4Biomedical by \citet{LLM4BiomedicalNLP} for BIOSSES (93.6), and SMART by \citet{jiang-etal-2020-smart} for STS (92.9). Models which BioLORD suprasses but remain worth mentioning include TLBNLP by \citet{peng-etal-2019-transfer} for MedSTS (84.8), BioNSTS by \citet{Blagec2019} for BIOSSES (81.9), and SiameseBERT by \citet{reimers-gurevych-2019-sentence} for SICK (74.5).
        }
    }
    \label{tab:results3}
\end{table*}

\begin{table*}
    \centering
    \begin{tabularx}{0.925\textwidth} { 
        >{\raggedright\arraybackslash}p{0.125\textwidth} 
        || >{\raggedright\arraybackslash}X
        | >{\raggedright\arraybackslash}X
        || >{\raggedright\arraybackslash}X 
        | >{\raggedright\arraybackslash}X 
        || >{\raggedright\arraybackslash}X 
        | >{\raggedright\arraybackslash}X 
	        }
         & \textbf{BioBERT} & \textbf{BioSyn} & \textbf{PubMedB.} & \textbf{SapBERT} & \textbf{STAMB2} & \textbf{BioLORD}\\
        \textbf{\textsc{MedNli-S}} & 89.0 & 89.5 & 90.1 & \textbf{90.5} & 89.4 & 89.9\\
    \end{tabularx}
    \caption{
        Accuracy on text similarity benchmark inspired from MedNLI. Higher is better.
    }
    \label{tab:results4}
\end{table*}

\begin{table*}
    \centering
    \begin{tabularx}{0.65\textwidth} { 
        >{\raggedright\arraybackslash}l 
        || >{\raggedright\arraybackslash}X
        || >{\raggedright\arraybackslash}X
        || >{\raggedright\arraybackslash}X 
	        }
         & \textbf{SapBERT} & \textbf{BioLORD} & \textbf{KRISSBERT} \\
        \textbf{NCBI-Diseases} & 63.0* & \textbf{68.4} & \underline{89.9}\\
        \textbf{MedMentions} & \textbf{37.6}* & 34.1 & \underline{70.7}\\
    \end{tabularx}
    \caption{
        Accuracy on named entity linking on MedMentions and NCBI. Higher is better. (*): Because we share the concerns raised by \citet{KRISSBERT}, and because we evaluate our model with the same methodology as theirs, we report in this table the results as presented in their paper, even for the SapBERT model.
    }
    \label{tab:results5}
\end{table*}

\begin{figure*}[]
    \centering
    \vspace{0.67cm}
    \textcolor{gray}{\textit{(Bold results indicate superiority amongst comparable models while \\underlined results indicate the overall state of the art)}}
    \vspace{0.66cm}
\end{figure*}

\subsection{Biomedical Entity Linking}

Another potential application of our model is Entity Linking. In the context of the biomedical domain, it consists in assigning to a given textual mention the biomedical concept which most faithfully represents it, among those defined in a target ontology. 

\textbf{NCBI-Diseases} \citep{Dogan2014-be} is an annotated corpus containing 6892 disease mentions, mapped to 790 unique diseases. 

\textbf{MedMentions} \citep{Mohan2019MedMentionsAL} is an annotated corpus aimed at the recognition of biomedical concepts in biomedical documents. Over 4,000 abstracts were annotated manually, for a total of over 350k linked mentions. These mentions are linked to their best match among the 3M+ concepts which were already referenced by UMLS in 2017. 

We finetune our model to make use of the context using the training set of MedMentions\footnote{We do so by forming positive pairs using the templates “\textit{[mention] [SEP] (context: [sentence])}” and  “\textit{[canonical-concept-name]}”. At inference time, we compute the representation for a mention with its context sentence between parentheses, then we identify the concept in the target ontology whose representation is the closest to that of the mention.}, and report our mapping accuracy in Table \ref{tab:results5}.

A variation of this task where our approach should particularly shine is entity normalization in non-exhaustive ontologies. Ontologies cannot possibly cover all concepts, but concepts absent from the ontology should ideally be normalized to broader, but semantically-compatible concepts.

\textbf{SCT-L2P} (our contribution; short for SnomedCT Leaf-to-Parent) is a benchmark in which concept names from leaf terms of the SnomedCT ontology are mapped within a reduced subset of the SnomedCT ontology, which is obtained by pruning all the leaf nodes from the hierarchy (to prevent self-mapping, see §\ref{section:ReproducingSCTL2P}).

In Table \ref{tab:results2a}, we report how often these leaf SnomedCT concepts were mapped to one of their parents, as opposed to a concept which is not one of their parents. This benchmark therefore evaluates whether the representations produced for biomedical concepts have a good hierarchical structure.

\subsection{Semantic Text Similarity}

Thanks to our initialization with a sentence embedding model, semantic text-similarity inference in the biomedical domain remains possible. We evaluate our performance on this task using 4 text-similarity benchmarks, after finetuning (see §\ref{section:ReproducingSTS}).

Text similarity benchmarks aim to evaluate how accurately models estimate the similarity in meaning between two pieces of text, usually sentences. As for the biomedical concept similarity task, human experts assign a similarity score to these pairs, and a set of such scores is averaged across experts to form a golden standard. We trained and evaluated all models on four such datasets.

\textbf{MedSTS} \citep{MedSTS} was developed for evaluating clinical semantic textual similarity. It contains 1,068 sentence pairs which were annotated by two medical experts with semantic similarity scores of 0-5 (low to high similarity). 

\textbf{BIOSSES} \citep{BIOSSES} considers small paragraphs rather than sentences, and focuses on scientific articles in the biomedical domain, rather than clinical notes. It is a challenging dataset because of the length of its entries.

\textbf{SICK} \citep{marelli-etal-2014-sick} consists of about 10k English sentence pairs, designed to be rich in lexical, syntactic, and semantic phenomena. Pairs have been annotated for relatedness on a 0-5 scale. 

\textbf{STS Benchmark} \citep{cer-etal-2017-semeval} regroups several other general-purpose text similarity datasets (and contains 8628 sentence pairs). 

Table \ref{tab:results3} reports the Pearson correlation between the similarity scores attributed to sentence pairs by the model under evaluation, and the gold standard for that sentence pair.

\subsection{Natural Language Inference}

Natural Language Inference (NLI) is the task of deciding whether a piece of evidence (usually a paragraph of text) can be used to support a given conclusion (usually a sentence about the same topic). The training and test data of NLI models usually contain pairs of sentences and a judgement on whether the first one, called the premise, entails (supports), contradicts (anti-supports), or is neutral towards a second sentence called the hypothesis.

\textbf{MedNLI} \citep{romanov-shivade-2018-lessons} is the dataset we used to perform this analysis. It was curated by doctors asked to provide three statements (one entailed, one contradicted, and one neutral) grounded in the medical history of a given patient. 

BioLORD is inherently not suited to capture the relation between two sentences, since it would encode them separately, unlike cross-encoder models that are typically trained for NLI tasks. 

Yet, we can hypothesize that semantically strong sentence-encoders should yield representations that are more similar for entailed sentence pairs than contradicted ones. We therefore propose the following evaluation strategy:
all premises of the input dataset for which both an entailed and a contradicted hypothesis have been curated are retained for usage as an input in triplet form. For these triplets, we report in Table \ref{tab:results4} whether the hypothesis most similar with the premise (in terms of cosine similarity) is indeed the entailed hypothesis, on average for the dataset.

\section{Discussion}
\label{section:Discussion}

Thanks to the experiments described in the previous section, the suitability of definition-based contrastive learning can now be demonstrated, which we will do in this section. We also discuss patterns appearing in our results, which might be useful in designing further improvements to our technique. 
	
Our pre-training strategy significantly improves the alignment with human judgement of biomedical entity representations (+27.6pts on MayoSRS compared to STAMB2, our initialization model; see Table \ref{tab:results1}). This improvement is more significant than those elicited by either SapBERT (+20.0pts) or BioSyn (+16.9pts). Since our model also performs better in absolute terms, this confirms that our pre-training strategy is the most effective.

Unlike both SapBERT and BioSyn, BioLORD's performance on general-purpose sentence representations remains competitive. It is however worth noting that the aforementioned pre-training strategies seem to impair general-purpose sentence representation, even after fine-tuning (see Table \ref{tab:results3}). Despite making use of descriptions and definitions in the training data, this remains true for BioLORD (-1.5pts on STS Benchmark). We dig deeper into these results in §\ref{section:DiscussSentences}.

In the specific case of biomedical sentences, the BioLORD pre-training nonetheless remains a net-positive. Thanks to this, we achieve state-of-the-art results for bi-encoders on MedSTS (86.3pts, +0.4pts over initialization), and we can report improvements over initialization on MedNLI-S as well (89.9pts, +0.5pts). 

\subsection{Biomedical concept representations}
\label{section:DiscussConcepts}

When it comes to the concept embedding task, BioLORD again outperforms its peers, achieving significantly higher Spearman correlation than SapBERT on all available benchmarks (+12.2pts for MayoSRS, +6.9pts for UMNSRS-R, +3.0pts for UMNSRS-S, +5.8pts for EHR-RelB; see Table \ref{tab:results1}).

BioLORD performs particularly well w.r.t.~its peers on the relatedness benchmarks. 
We attribute this to the fact our model is trained on all the relationship types described in UMLS: its representations might therefore encompass more ways for concepts to be related to each other, compared to models trained on the hierarchical relationships only. This matters less for similarity benchmarks, because similar entities are more likely to be close in the hierarchical graph, too.

Our improved representations provide additional benefits, as shown by our SCT-L2P evaluation task. When matching leaf-concepts against a leaves-removed ontology, BioLORD was able to match leaf-concepts to their parent concepts more often than SapBERT (+7.9pts on SCT-L2P Accuracy@1, see Table \ref{tab:results2a}). BioLORD also assigns significantly fewer concepts to a representation so disconnected from those of its parents that none appear in the 1000 nearest neighbors (3.8\% vs 8.6\%). BioLORD therefore achieves a much higher mean-reciprocal rank than SapBERT (+9.3pts, see Table \ref{tab:results2a}).

It is worth noting that combining \citep{Mao2020-ov} or enhancing \citep{KalyanHybridEmbeddings} concept-name representations with graph embeddings remains the state of the art in some benchmarks even after our contributions, which means that further improvements might still be achievable by merging BioLORD representations with graph-based representations. We leave this investigation for a future article.

For completeness, we also evaluated our model on the more traditional benchmarks of biomedical entity linking. As our results on MedMentions and NCBI-Diseases show, BioLORD remains competitive with other alternatives on this task as well (see Table \ref{tab:results5}).

\subsection{Biomedical sentence representations}
\label{section:DiscussSentences}

Unlike previous models, BioLORD also aims to provide good representations for biomedical sentences from clinical notes, and indeed achieves state-of-the-art results for MedSTS among bi-encoder models (see Table \ref{tab:results3}). 

While cross-encoder models are known to achieve better results, bi-encoder models remain important for search and retrieval scenarios as they support the independent treatment of documents, enabling parallel workflows. Given that most hospitals accumulate millions of clinical notes, this is a highly desirable feature.

What makes our model particularly suitable for clinical note understanding is that it shows solid performance on general-purpose text as well, strongly outperforming all the other state-of-the-art biomedical models in this setup. This is true on both the SICK and STS Benchmark datasets (+9pts over SapBert on SICK, +5.5pts on STS Benchmark). Because clinical notes often contain a mix of text in medical jargon and more mundane discourse, we believe using models generalizing well to multiple domains is valuable, as both types of discourse are likely to carry useful information. 

\textbf{Benefit of the descriptions:} As we expected, the performance degradation on SICK (-1.4pts) remained smaller than on STS Benchmark (-3.5pts) in the case of BioLORD, while the SapBERT pre-training caused significant performance reduction on SICK (-6pts). We can explain this by the way the SICK dataset is constructed. Unlike most STS benchmarks, SICK tries to require as little as possible the understanding of a domain language (e.g. idiomatic multiword expressions, named entities, etc...) while putting more attention on lexical, syntactic and semantic phenomena that generalize well across domains (for instance, by replacing nouns or adjectives by equivalent relative clauses). This mixes well with our pre-training approach which combines nouns and descriptions. 

In the case of BioLORD, a large part of the performance degradation on general-purpose semantic text similarity performance (the STS Benchmark) is therefore likely caused by the loss of knowledge about concepts unrelated to the biomedical domain. 
This is reassuring, because that type of knowledge is less likely to be useful in biomedical text understanding than the general understanding of sentence structures which is tested by SICK.

\section{Limitations}

While BioLORD models perform better on general-purpose text than other state-of-the-art biomedical models, we would like to recognize that no guarantee can be made regarding their ability to make sense of non-ontological modalities of languages in the biomedical domain (spoken language, social media, online forums, etc…). This might cause issues in circumstances where the content being processed is not originating from clinicians.

BioLORD models are not well suited to tasks requiring the latest scientific knowledge, for example found in PubMed papers, because the origin of most of their biomedical knowledge is ontologies. This type of knowledge might be better encoded in PubMedBERT-based models. Hybrid approaches might therefore be worthwhile for natural language inference and scientific paper analysis scenarios.

Another limitation worth mentioning is that, unlike Graph Neural Networks, BioLORD models do not provide a single and unique embedding for each concept of an ontology, because concepts might have more than one known name. As representations are produced for names rather than concepts, each name will produce a different (but hopefully close) representation for a concept. This might amper rule-based decision processes.

Despite these limitations, we envision that most use cases traditionally considered in the biomedical domain are suitable for the usage of BioLORD models, in an end-to-end setup.

\vspace{0.27cm}

\section{Conclusion}

The results detailed in this paper demonstrate that our name-to-definition contrastive pre-training strategy (BioLORD) is able to produce bi-encoder models with a state-of-the-art performance for both biomedical concept similarity and clinical sentence similarity, using one set of weights. 

As part of our investigation, we were also able to show that appropriate finetuning strategies can enable general-purpose models (like STAMB2) to fully overcome their initial disadvantage at bio-medical text understanding compared to models of the same size trained on biomedical data only (such as PubMedBERT). This can be achieved while preserving most of their pre-existing general-purpose text understanding capabilities.

By releasing a bi-encoder model able to embed both clinical sentences and biomedical concepts in the same latent space, we immensely simplify 
some clinical language processing scenarios, like clinical note retrieval and diffuse information extraction. For this reason, we hope that future works in the biomedical domain will continue to combine clinical sentences and biomedical concept benchmarks to produce directly usable end-to-end models. 

\raggedbottom

\section{Ethical considerations}

Healthcare is a very sensitive domain, and we could not simply conclude this article without first taking some time to reflect on the ethical implications of the usage in the real world of models trained using our BioLORD methodology. 

Using machine learning models to affect a patient's treatment is never a decision to be taken lightly, and we are conscious of the fact medical practitioners should always be kept in the loop in such a process. Machine learning models might help surface information more easily, but should never replace medical practitioners entirely, given the need for verification of the information extracted by models, and for guiding the usage that is made of the information once extracted. 

Because of our intentional focus on publicly-available ontologies and annotated benchmarks, only a limited set of sources were used to train our models. However, we are aware that knowledge originating only from ontologies might need to be supplemented by other data sources to provide trustworthy results in real-word settings. We would therefore like to urge machine learning specialists to consider the issue of human-machine alignment before using models built using our approach.

In-house data and annotations will be crucial to achieve good outcomes in the real world. We hope hospitals will continue to invest in the annotation of data to fully realize the potential of machine learning models, but we recognize this is a time-consuming process. Better tools would probably help medical experts to spend less time on data annotation than they need to spend today. 

We also want to point out that multi-linguality is important for hospitals taking care of patients coming from regions where different languages are spoken. End-to-end models are very suitable for cross-lingual distillation, which require less data than training from scratch \citep{reimers-gurevych-2020-making}. This justifies our decision to train an end-to-end model, and we aim to talk more about multi-linguality in the near future. 

Taking the previous points in consideration leads us to believe that the publication of this model and of our results can have a positive impact on society in addition to the machine learning field, but we remain available to discuss further ethical concerns.

\newpage

	\bibliography{anthology,custom}
\bibliographystyle{acl_natbib}

\newpage
\appendix
\renewcommand\thefigure{\thesection.\arabic{figure}}    

\section{Replication details}
\label{section:ReproducingDetails}
\setcounter{figure}{0}

Given this article covers many benchmarks and datasets, it is not practical to provide all the details required for reproduction in the main text. Details which we do not believe contribute to the understanding of our paper are listed here.

\subsection{BioLORD Dataset}
\label{section:ReproducingBioLORD}

The dataset contains 100M pairs (86M with descriptions, 14M with definitions). Another set of 20M descriptions based on the same knowledge graph serves as a development set (86M generations certainly do not exhaust the graph). However, this would not be a suitable test set. Instead, a \textit{test of time} consisting of new concepts currently absent from UMLS would make more sense, but this will have to wait until enough new concepts have been added to UMLS. Our supplementary materials contain a sample of our train set, and we uploaded our full train data on huggingface (24Gb file).

\subsection{BioLORD Pre-training}
\label{section:ReproducingPreTraining}

Given the time required for training BioLORD models with the hardware available to us, only cursory hyperparameter tuning was performed. Our investigation focused on the type of loss (InfoNCE vs MegaBatchMarginLoss), the ratio of definition and description, the number of training examples (20M vs 100M), and the batch size (64 vs 96). We kept all other hyperparameters at their default value (\textit{AdamW} for the optimizer, \textit{WarmupLinear} for the scheduler, \textit{2e-5} for the learning rate, \textit{5\% of the data} for the warmup window, \textit{0.01} for the weight decay, \textit{1} for the number of epochs, \textit{PyTorch 1.7.1 AMP} for the mixed-precision training). 

Rather than using the downstream task as evaluation, a manual investigation was carried out by inspecting the latent space neighborhood of a set of biomedical concepts\footnote{cat / cat scan / cat scratch / cat scratch disease / cat scratch disease antigen / nephrectomy / partial nephrectomy / lobotomy / hand / left hand / cleft hand / apyrexial / afebrile / ranitidine / ranitidine allergy / falcons / belgium / apyrexial / afebrile / hepatic arteriogram / bowling / bowling alley / bowling alley manager / endometriosis / heparin sodium}. Due to the subjective nature of this test, we cannot discard the chance that another set of hyperparameters might have yielded to a better model.

The model whose results are described in this paper (100M pairs (15\% of definitions and 85\% of descriptions) in batches of 64 pairs) took 7 days of training on a NVIDIA V100 GPU.

\subsection{SnomedCT-L2P Dataset}
\label{section:ReproducingSCTL2P}

SnomedCT contains about 700k concept names, among which about 550k concern leaf concepts, and about 150k concern non-leaf concepts. These concepts have hierarchical relationships defined between each other, and the goal of this dataset is to detect whether leaf concepts have a representation which is closer to the representation of their hierarchical parents than to the representation of other concepts in the ontology. We do not define a train / dev split, and use the entire dataset as a test set, because we intend to evaluate model performance without further finetuning. SnomedCT is an ever-evolving dataset, so our supplementary materials contain the files we used for evaluation, so that our results can be replicated.

\subsection{STS Finetuning}
\label{section:ReproducingSTS}

For all the finetuning experiments, the following hyperparameters have been selected: \textit{AdamW} for the optimizer, \textit{WarmupLinear} for the scheduler, \textit{6e-6} for the learning rate, \textit{5\% of the data} for the warmup window, \textit{0.01} for the weight decay, \textit{10} for the number of epochs (the weights are saved after every epoch, and the best weights for the dev set are retained for the final evaluation), \textit{64} for the batch size, \textit{PyTorch 1.7.1 AMP} for the mixed-precision training. The finetuning time varies per dataset, but at most reaches a couple of minutes (same hardware as before).

\subsection{MedMentions Finetuning}
\label{section:ReproducingMedMentions}

For all the finetuning experiments, the following hyperparameters have been selected: \textit{AdamW} for the optimizer, \textit{WarmupLinear} for the scheduler, \textit{2e-5} for the learning rate, \textit{5\% of the data} for the warmup window, \textit{0.01} for the weight decay, \textit{1} for the number of epochs, \textit{64} for the batch size, \textit{PyTorch 1.7.1 AMP} for the mixed-precision training. The finetuning requires about 4.5 hours to finish (same hardware as before).

\subsection{Downloading our model}
\label{section:DownloadingOurModel}

Our biomedical model is hosted on \href{https://huggingface.co/FremyCompany/BioLORD-STAMB2-v1}{\nobreak{huggingface.co/} \nobreak{FremyCompany/BioLORD-STAMB2-v1}}. \\
\\
Its dataset is hosted on: \href{https://huggingface.co/datasets/FremyCompany/BioLORD-DS}{\nobreak{huggingface.co/datasets/} \nobreak{FremyCompany/BioLORD-DS}}.\\
\\
Our supplementary materials include our training code and a sample of our dataset.

\newpage
\section{Qualitative analysis}
\label{section:QualitativeAnalysis}
\setcounter{figure}{0}

In this appendix, we compare the similarity matrix of a few biomedical concepts, as generated by BioSyn, SapBERT, and BioLORD. This highlights the improvements of the latter over its peers.

Within each group of three, concepts are significantly more related to each other than the rest.

\begin{figure}[h]
    \centering
    \includegraphics[width=\linewidth]{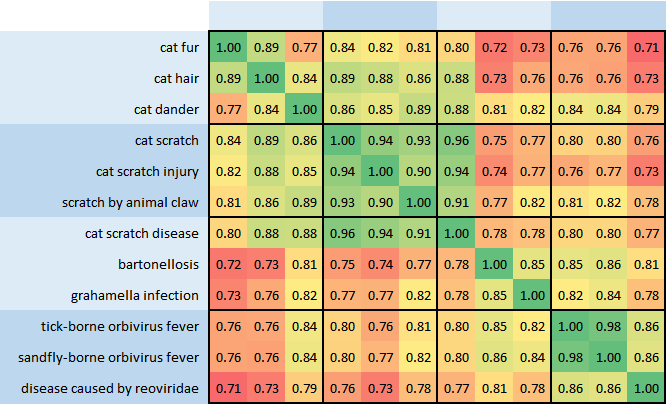}
    \caption{Concept Similarity matrix for BioSyn}
    \label{fig:SimilarityMatrixBioSyn}
\end{figure}

\begin{figure}[h]
    \centering
    \includegraphics[width=\linewidth]{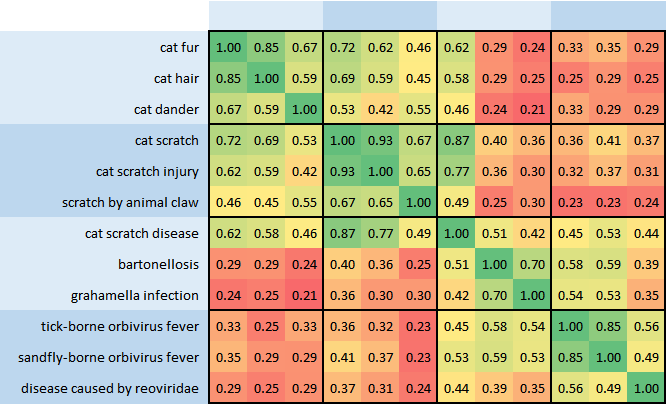}
    \caption{Concept Similarity matrix for SapBERT}
    \label{fig:SimilarityMatrixSapBERT}
\end{figure}

\begin{figure}[h]
    \centering
    \includegraphics[width=\linewidth]{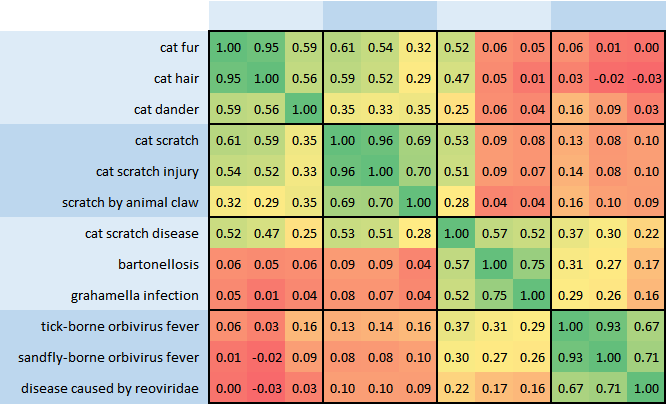}
    \caption{Concept Similarity matrix for BioLORD}
    \label{fig:SimilarityMatrixBioLORD}
\end{figure}

As can be seen above, BioLORD does a much better job clustering the related concepts together, without mixing them with the other groups. It even clusters "cat scratch disease" with other types of Bartonella infections.

\newpage
\section{Acknowledgments}
\label{section:Acknowledgments}
\setcounter{figure}{0}

This work would not have been possible without the joint financial support of the Vlaams Agentschap Innoveren \& Ondernemen (VLAIO) and the RADar innovation center of the AZ Delta hospital group. 

\vspace{0.3cm}

I am especially indebted to Ir. Peter De Jeager, Chief Innovation Officer of RADar, for his role in setting up and keeping on track the Advanced Data-Aided Medicine project (ADAM), and ensuring prompt and smooth communication between me and the other shareholders working at the hospital.

\vspace{0.3cm}

A particular thanks is also extended to the team maintaining and developing GPULab, the machine learning infrastructure for AI computing built in collaboration between UGent, UAntwerpen and the Imec research and development center.

\vspace{0.3cm}

Finally, I would also like to thank my co-supervisors, Kris Demuynck and Thomas Demeester, for their support and constructive advice during the ideation process, and all along the development of this project up to this very article.

\end{document}